\documentclass[11pt]{article}

\usepackage[preprint]{acl}

\usepackage{times}
\usepackage{latexsym}

\usepackage[T1]{fontenc}

\usepackage[utf8]{inputenc}

\usepackage{microtype}

\usepackage{inconsolata}

\usepackage{graphicx}

\usepackage{amsmath}
\usepackage{xcolor}
\usepackage{booktabs}
\usepackage{multirow}
\usepackage{array}
\usepackage{amssymb}
\usepackage{bbm}

\newcommand{\ourmetric}{\textsc{Vital}}
\newcommand{\vitalprecision}{\ourmetric{}$_\textsc{Prec.}$}
\newcommand{\vitalrecall}{\ourmetric{}$_\textsc{Rec.}$}
\newcommand{\responselevelprecision}{\ourmetric{}$_\textsc{RLP}$}
\newcommand{\responselevelrecall}{\ourmetric{}$_\textsc{RLR}$}
\newcommand{\nuggetrecall}{\textsc{NuggetRecall}}
\newcommand{\ourdataset}{\textsc{VitalErrors}}

\newcommand{\factscore}{\textsc{FActScore}}
\newcommand{\veriscore}{\textsc{VeriScore}}
\newcommand{\safe}{\textsc{Safe}}
\newcommand{\bright}{\textsc{BRIGHT}}

\newcommand{\hotpotqa}{\textsc{HotpotQA}}
\newcommand{\triviaqa}{\textsc{TriviaQA}}
\newcommand{\naturalquestions}{\textsc{NQ}}

\title{All Claims Are Equal, but Some Claims Are More Equal Than Others: \\
Importance-Sensitive Factuality Evaluation of LLM Generations}

\author{
 \textbf{Miriam Wanner\textsuperscript{1*}},
 \textbf{Leif Azzopardi\textsuperscript{2, 3}},
 \textbf{Paul Thomas\textsuperscript{2}},
 \textbf{Soham Dan\textsuperscript{2}},
\\
 \textbf{Benjamin Van Durme\textsuperscript{1,2}},
 \textbf{Nick Craswell\textsuperscript{2}}
\\
 \textsuperscript{1}Johns Hopkins University,
 \textsuperscript{2}Microsoft,
 \textsuperscript{3}University of Strathclyde,
\\
 \href{mailto:mwanner5@jhu.edu}{\texttt{mwanner5@jhu.edu}}
}

\begin{document}
\maketitle
{
\def\thefootnote{*}\footnotetext{Work done during an internship at Microsoft.}
}

\begin{abstract}

 Existing methods for evaluating the factuality of large language model (LLM) responses treat all claims as equally important. 
 This results in misleading evaluations when vital information is missing or incorrect as it receives the same weight as peripheral details, raising the question: how can we reliably detect such differences when there are errors in key information? Current approaches that measure factuality tend to be insensitive to omitted or false key information. 
 To investigate this lack of sensitivity, we construct \ourdataset{}, a benchmark of 6,733 queries with minimally altered LLM responses designed to omit or falsify key information. Using this dataset, we demonstrate the insensitivities of existing evaluation metrics to key information errors. 
 To address this gap, we introduce \ourmetric{}, a set of metrics that provide greater sensitivity in measuring the factuality of responses by incorporating the relevance and importance of claims with respect to the query.
 Our analysis demonstrates that \ourmetric{} metrics more reliably detect errors in key information than previous methods. Our dataset, metrics, and analysis provide a foundation for more accurate and robust assessment of LLM factuality.

\end{abstract}




\section{Introduction}

\begin{figure}
    \centering
    \includegraphics[width=1\linewidth]{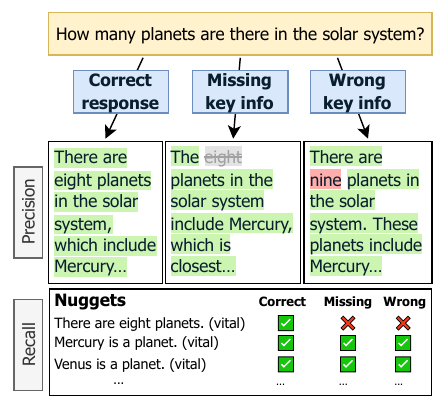}
    \caption{Three responses to the same query receive similarly high precision and recall scores. Precision and recall metrics are insensitive to errors in key information for long responses. Both erroneous responses still achieve high precision and recall scores.}
    \label{fig:motivation}
\end{figure}

\begin{figure*}
    \centering
    \includegraphics[width=1\linewidth]{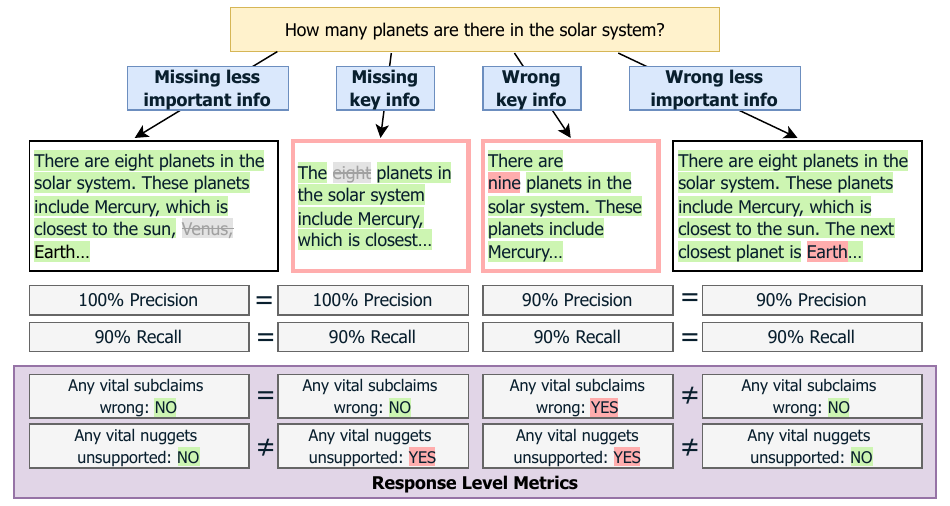}
    \caption{Four responses to the same query with varying error type and severity, demonstrating that not all errors are equal. Standard precision/recall give similar scores, while response-level metrics distinguish missing or wrong vital information from peripheral errors.}
    \label{fig:motivation-unequal}
\end{figure*}

Fact checking machine generated text has become increasingly important with the growing mainstream use of Large Language Models (LLMs). 
These models are known to produce hallucinations, which can sometimes be harmful \citep{bender2021dangers, dev-etal-2022-measures}. 
As a result, fact checking is often incorporated into such systems, to provide higher quality responses by removing such errors.
Previous methods for evaluating factual precision employ a \textit{decompose-then-verify} framework, which involves first decomposing claims from generated responses into atomic subclaims\footnote{In this work we use the term \textit{subclaim}, but other works use atomic-fact, -claim, or -preposition for the same concept.}, independently verifying each, and aggregating into a score \citep{min-etal-2023-factscore, song-etal-2024-veriscore, wanner-etal-2024-closer, huang2025medscore}.
Similar methods are used in evaluating recall, where references are used to generate \textit{nuggets}--pieces of information vital for answering a given query. 
The percentage of nuggets supported by the response is the recall score \citep{voorhees-2003-nuggets, pradeep2024autonuggetizer, pradeep2025greatnuggetrecallautomating}. 

These factuality metrics treat each piece of information (subclaim) equally, and tend not to explicitly consider whether a claim in the response is vital to answering a user's query. 
Not all parts of the response are equally relevant or important, and as such, some factuality-based errors are more egregious than others. 
However, current metrics are unable to distinguish or fail to delineate between similar responses with specific errors.
One missing or wrong key piece of information can invalidate an otherwise good response, which should be reflected in evaluations.
Figure \ref{fig:motivation} shows three different responses to a query asking for the number of planets in the solar system.  
The first response correctly answers with ``eight,'' the second provides the list of planets, but is missing the key piece of information (``eight''), and the third provides the incorrect answer ``nine.'' 
Despite the missing or incorrect key information, both erroneous responses still achieve high precision and recall scores. 
The most important piece of information (the answer ``eight'') is treated as only one small piece of information in a long response, where the final score is agnostic to its importance in fundamentally answering the user's query.

In this paper, we investigate the impact of \textit{small} but \textit{important} errors due to missing key or incorrect claims in responses that are vital to answering the query, whether these can be detected using traditional fact checking metrics, and if considering importance improves their detectability and sensitivity.
We introduce a set of \ourmetric{} metrics for identifying errors in key information. 
We then construct a new dataset that allows us to rigorously evaluate how key information errors affect the scoring of these systems. 
To create our dataset, we adversarially alter machine generate responses to queries, and generate minimally changed responses with missing and wrong key information to create the \ourdataset{} dataset.
We run previous methods of precision and recall, finding they are insensitive to these key errors, and show that \ourmetric{} metrics are sensitive to these key errors, reflecting when key information is missing or wrong. 

Our contributions are as follows: (1) We demonstrate how existing methods fail to capture important differences in responses. We show that there is no notion of importance within subclaims, and why this is a problem. (2) We introduce a new set of metrics, which we call \ourmetric{}, which includes subclaim ranking and importance labeling for precision metrics. (3) We construct \ourdataset{}, a challenging dataset for evaluating key information errors. Using queries from six QA datasets, we generate normal LLM responses, and adversarially alter each response to omit or falsify key information. We demonstrate better error detection using \ourmetric{} metrics on the \ourdataset{} benchmark.





\section{Metrics}\label{sec:metrics}

To ground our proposed \ourmetric{} metrics, we first provide an overview of existing factuality and nugget based metric frameworks.

\begin{table*}[ht]
\centering
\small
\begin{tabular}{m{2.2cm}m{2.5cm}m{2cm}m{1cm}m{5.2cm}}
\toprule
\textbf{Data Type} & \textbf{Dataset} & \textbf{Subsets} & \textbf{Size} & \textbf{Description} \\ 
\hline
\multirow{6}{*}{Open-Ended} & \factscore{} Bios & - & 500 & Generated people biographies and corresponding Wiki page \\ 
\cline{2-5}
& \textsc{WildHal.} & Cult. \& Ent., Geographic & 2484 & Entities from user conversations and curated web page knowledge source \\
\cline{2-5}
& \bright{} & Bio, E.S., Econ, Psych, Rob., S.O., S.L. & 749 & Real-world queries from StackExchange and linked sources\\
\hline
\multirow{6}{*}{Single-Answer} & \hotpotqa{} & - & 1000 (subset) & Multi-hop reasoning questions and supporting documents\\
\cline{2-5}
& \naturalquestions{} & - & 1000 (subset) & Real Google search queries and results\\
\cline{2-5}
& \triviaqa{} & - & 1000 (subset) & Trivia questions and reference documents\\
\bottomrule
\end{tabular}
\caption{Overview of datasets used to construct \ourdataset{}, covering open-ended and single-answer queries across six sources, totaling 6,733 instances.}\label{tab:datasets}
\end{table*}

\subsection{Factual Precision}
\citet{min-etal-2023-factscore} introduced a framework for fact verification. The system first decomposes LLM-generated text into ``\textit{atomic facts}'', which are then independently verified against a trusted grounding source.
Many variants of this decompose-then-verify framework exist. While previous works focused on factual \textit{precision}, \safe{} \citep{wei2024safe} extends this work to include recall, with a selected hyperparameter $\mathit{K}$, the number of supported facts for a response to achieve full recall. 
\veriscore{} \citep{song-etal-2024-veriscore} builds on \safe{}, using the same measure of recall, while only decomposing text into \textit{verifiable} subclaims. \citet{liu2025verifactenhancinglongformfactuality} highlight the issue of recall in previous metrics and introduce \textsc{VeriFact}, evaluating responses with \textit{reference fact sets}, much like nuggets. \citet{zhao-2025-response-length} study the length-factuality tradeoff, finding a quality degradation with increased response length.

\citet{wanner-etal-2024-closer} examined the decomposition step, showing these factual precision pipelines are sensitive to different decomposition methods. \citet{jiang2024core} extended this work to fix decomposition redundancy, and introduce \textsc{Core}, a method for subclaim selection based on uniqueness and informativeness. \textsc{Core} is a subclaim filtering method, where as this paper works on evaluating and detecting errors in these important subclaims.
\citet{gunjal-durrett-2024-molecular} and \citet{wanner2024dndscore} formalized, implemented, and incorporated decontextualization in factual precision frameworks. Further works have researched diverse factual information \citep{samarinas2025factualaccuracy}, better localizing factual errors \citep{cattan-2024-localizing-factual}, entailment reasoning \citep{eliav-2025-clatter}, and factual precision in medicine \citep{huang2025medscore}. \citet{godbole-jia-2025-pitfalls} re-evaluated these evaluation metrics, and find pitfalls in current factuality metrics. 
We use \factscore{} in our experiments for factual precision evaluation, employing \texttt{gpt-4o} for atomic fact generation and evaluation.


\subsection{Nugget Recall}

\citet{voorhees-2003-nuggets} introduced \textit{nuggets} as a minimal, atomic claim that sounds as a correct and useful piece of information in answering a query. 
This formalization provided a foundation for evaluating relevance and recall of retrieved documents or responses for a given query. 
Each nugget is assigned an importance label, vital or okay, depending on how important the nugget is for answering the query. \citet{lin-2006-pyramid-nuggets} showed how importance can be extended to grades of importance by combing ratings from multiple assessors (i.e., nugget importance is based on annotator popularity).
Nugget based evaluation metrics have been used for not only assessing answer quality, but also document relevance, summarization \citep{nenkova-2004-pyramid}, RAG \citep{pradeep2025greatnuggetrecallautomating}, and conversational search \citep{abbasiantaeb2025conversational}.

Recently, \citet{pradeep2025greatnuggetrecallautomating} implement an automated LLM-based framework using the original vital/okay labels, called AutoNuggetizer, for nugget creation, ranking, assignment, and scoring \citep{pradeep2024autonuggetizer}. Given a list of nuggets, we can then determine which nuggets are most important to the user.
We use AutoNuggetizer with \texttt{gpt-4o} as the LLM for generating nuggets, importance labeling, and evaluating responses. A query and set of documents are provided to the LLM, and a set of nuggets are created. The LLM is then prompted to label each nugget as either \textit{vital} (needed in a response) or \textit{okay} (good to have, but not necessary). Finally, the LLM is given the set of nuggets and response to the query, and evaluates if each nugget is supported, partially supported, or not supported by the response. \citet{pradeep2025greatnuggetrecallautomating} report multiple metrics for recall evaluation using different combinations and weighting of vital/okay nuggets, which can be supported, partially supported, or not supported. For this paper we use their ``All Strict'' metric, which we denote as \nuggetrecall{}.

\subsection{Comparison of Metrics}

A central challenge in evaluating text lies in how we decompose information into units that can be assessed or used for assessment.
Nuggets \citep{voorhees-2003-nuggets} and claim decomposition \citep{min-etal-2023-factscore, wanner-etal-2024-closer} share surface similarities, but differ in scope, assumptions, and grounding. 
Here we lay out the differences, which is important to understand the contribution of our work.

The nugget framework assumes that a sufficiently broad pool of potentially relevant material can be assembled--typically by pooling results from multiple retrieval systems--and that annotators (or now LLMs) can then extract nuggets representing the key information needs of a query. Thus, nugget-based evaluation is inherently \textit{per-query} and grounded in relevance judgments across multiple sources. 
In contrast, factuality metrics operate by decomposing an individual response into smaller \textit{per-response} units, or subclaims. These subclaims are then verified against retrieved evidence. 
This framework does not rely on pooling across systems, but rather assumes that the response itself contains the material to be fact-checked. 
Subclaims emphasize verification rather than relevance, and their validity is tightly coupled to the retrieval system used for grounding.

Both frameworks share the intuition of breaking down long-form text into smaller, verifiable units, however they diverge in scope (query-centric vs. response-centric), evidence base (rich pool of candidate material vs. retrieval material), objective (relevance vs. factual correctness), and granularity (nugget vs. claim). 
In this work, we build on these previous systems. We adopt the nugget notion of \textit{importance}, i.e., that some pieces of information are vital to user needs, while retaining the original method of claim decomposition. These metric changes address issues of equal subclaim weighting and lack of subclaim importance.






\section{Methods}


To further motivate the need for more sensitive evaluation metrics, consider the example in Figure \ref{fig:motivation-unequal} -- which shows that not all errors are equal.
On the left are two responses with missing information, one which omits the answer ``eight planets,'' the other which omits ``Venus is a planet.'' 
Both answers achieve the same high precision and recall scores, despite varying quality. 
Similarly, the two responses on the right both contain wrong information, one answers the query wrong with ``nine planets,'' the other mixing up Earth and Venus in planet ordering. 
As with the previous two examples, both responses have equally high scores. 
Incorrect or missing key information is more harmful than incorrect or missing peripheral information, since this key information receives the most attention. 
Comparing these responses only using these metrics, they all seem to be of similar quality, despite varying magnitude of error.
On the other hand, using response-level metrics (shown in the purple box) allows us to differentiate based on errors in key information, correctly flagging the middle responses as missing and/or containing false key information.
We blend the notion of importance in nuggets into \factscore{} by introducing \ourmetric{} metrics and construct \ourdataset{} in the following sections. Put simply, not all claims are equal; some are more important than others.

\begin{table*}[ht]
\centering
\small
\begin{tabular}{p{1.2cm}p{14cm}}
\toprule
\textbf{Type} & \textbf{Response} \\
\midrule
\multirow{1}{*}{\textbf{Query}} & In a paragraph, could you answer: Of which tribe of Red Indians was Geronimo a chief \\
\midrule
\multirow{5}{*}{\textbf{Normal}} & Geronimo was a prominent leader and medicine man of the Apache tribe, specifically belonging to the Bedonkohe band of the Chiricahua Apache. Although he is often referred to as a chief, Geronimo was not a hereditary chief in the traditional sense but rather a respected warrior and spiritual figure who played a crucial role in resisting U.S. and Mexican military campaigns during the late 19th century. His leadership and fierce resistance made him a symbol of Native American defiance and resilience during a time of intense conflict and displacement. \\
\midrule
\multirow{4}{*}{\textbf{Missing}} & Although he is often referred to as a chief, Geronimo was not a hereditary chief in the traditional sense but rather a respected warrior and spiritual figure who played a crucial role in resisting U.S. and Mexican military campaigns during the late 19th century. His leadership and fierce resistance made him a symbol of Native American defiance and resilience during a time of intense conflict and displacement. \\
\midrule
\multirow{5}{*}{\textbf{Wrong}} & Geronimo was a prominent leader and medicine man of the Sioux tribe, specifically belonging to the Bedonkohe band of the Chiricahua Apache. Although he is often referred to as a chief, Geronimo was not a hereditary chief in the traditional sense but rather a respected warrior and spiritual figure who played a crucial role in resisting U.S. and Mexican military campaigns during the late 19th century. His leadership and fierce resistance made him a symbol of Native American defiance and resilience during a time of intense conflict and displacement. \\
\bottomrule
\end{tabular}
\caption{An instance from \ourdataset{}, with a query from \triviaqa{}, and three response types. The normal response identifies Geronimo as Apache, while missing and wrong variants omit or replace this key claim.}
\label{tab:data-examples}
\end{table*}

\subsection{\ourmetric{} Metrics} 

We draw upon \factscore{} and nuggets to propose a novel importance based factuality score of key information for evaluating responses.
Using a set of decomposed subclaims, we then rank claims by \textit{query importance}. 
A claim exhibits high query importance when it addresses a central aspect of the query, and low query importance when it contributes only peripheral or background information. 
We task an LLM to rank subclaims independent of correctness, instead only with respect to the user's query. 
In this same prompt, we also ask the LLM to give importance labels to each subclaim. 
We ask for an importance label to be assigned to each subclaim: \textit{vital} (and thus critically important in answering the user's query), \textit{okay} (somewhat important), and \textit{less important}.
From this, we have a set of subclaims/nuggets, labeled and ordered by importance. 
Prompt and model details can be found in appendix section \ref{app:prompts}.

Given the labels and ranking, we present two ways of evaluating key information. 
We evaluate using subclaims in precision metrics and nuggets in recall metrics. 
First we have decomposition level evaluations: \vitalprecision{} and \vitalrecall{}. 
We only report the precision and recall for the vital set of subclaims/nuggets. 
This gives us a notion for how precise the response is and the coverage for core pieces of information. 
Let $\hat{V} = \{\hat{v}_{1}...\hat{v}_{n}\}$ be the set of vital subclaims from a response and $V = \{v_{1}...v_{n}\}$ be the set of vital nuggets for the query used to generate the response. 
Then we have 
\begin{align*}
    \text{\vitalprecision{}} &= \frac{|\hat{v}_i\text{ supported}|}{|\hat{V}|}\\
    \text{\vitalrecall{}} &= \frac{|v_i\text{ supported}|}{|V|}
\end{align*}
where subclaim support is judged against trusted sources, and nugget support is judged against the response.

Then we have response-level evaluations: \responselevelprecision{} (Response-Level Precision) and \responselevelrecall{} (Response-Level Recall). Each are boolean evaluating if any of the vital subclaims are wrong (for \responselevelprecision{}) and if any of the vital nuggets are missing (for \responselevelrecall{}). 
\begin{align*}
    \text{\responselevelprecision{}} &= \begin{cases}
        1, \text{any }\hat{v_i}\text{ unsupported} \\
        0, \text{otherwise}
    \end{cases}\\
    \text{\responselevelrecall{}} &= \begin{cases}
        1, \text{any }v_i\text{ unsupported} \\
        0, \text{otherwise}
    \end{cases}
\end{align*}
These metrics focus on only \textit{vital} information, and evaluate and detect errors only in this part of the response\footnote{We look into a weighted metric using \textit{all} subclaims, described in Appendix \ref{appendix:results}.}.


\subsection{\ourdataset{} Dataset}

To explore how current metrics and our proposed metric perform with respect to different types of errors, we constructed an evaluation test set, where we perturb LLM responses to omit or falsify key information.

\paragraph{Queries}
We used a total of 6,733 queries from six datasets of varying query type. 
We used data with \textit{open-ended} and \textit{single-answer} queries. 
Open-ended queries have little constraints and many possible answers are correct (e.g., ``Tell me about George Washington''). 
Single-answer queries have a particular piece of information that must be included in a correct answer (e.g., ``How many planets are in the solar system?''). 
Table \ref{tab:datasets} shows a summary of the datasets we used. These consisted of commonly used datasets focused on: entities \citep{min-etal-2023-factscore, zhao2024wildhallucinations}, QA \citep{yang2018hotpotqa, kwiatkowski2019naturalquestions, 2017triviaqa}, and complex reasoning \citep{su2024BRIGHT}. 
We prompted an LLM with the queries in these datasets to get long-form machine generated answers. 
We then generated a \textit{missing} and \textit{wrong} adversarial version of each response.

\paragraph{Adversarial \textit{Missing} and \textit{Wrong} Responses}
Given a query and a response, we can adversarially perturb the response to omit or falsify key information, while minimally altering the rest of the response. To do this we prompt an LLM with query and response pairs, and instruct the model to remove or alter a single piece of information that is most important for answering the query. For every query, we then have three responses: (1) \textit{normal}, (2) \textit{missing}, where key information is omitted, and (3) \textit{wrong}, where key information is incorrect. We can evaluate these responses with the metrics outlined in the following sections to assess how metrics shift for these minimal changes with a large impact on answer quality.

\begin{table}[!t]
  \centering
  \begin{tabular}{llrrr}
    \hline
    \textbf{   } & \textbf{Metric} & \textbf{Normal} & \textbf{Missing} & \textbf{Wrong} \\
    \hline
    \multicolumn{5}{l}{\textbf{Open-Ended}} \\
    \multicolumn{5}{l}{\textit{Subclaims}} \\
    & vital & 11.34 & 8.13 & 11.66 \\
    & okay & 11.37 & 7.60 & 10.96 \\
    & less & 6.34 & 4.23 & 6.47 \\
    & \textit{TOTAL} & \textit{29.04} & \textit{19.96} & \textit{29.09} \\
    \multicolumn{5}{l}{\textit{Nuggets (response independent)}} \\
    & vital & \multicolumn{3}{c}{9.84} \\
    & okay & \multicolumn{3}{c}{14.37} \\
    & \textit{TOTAL} & \multicolumn{3}{c}{\textit{24.21}} \\
    \hline
    \multicolumn{5}{l}{\textbf{Single-Answer}} \\
    \multicolumn{5}{l}{\textit{Subclaims}} \\
    & vital & 4.31 & 3.59 & 4.64 \\
    & okay & 7.69 & 6.49 & 7.81 \\
    & less & 10.36 & 6.83 & 9.84 \\
    & \textit{TOTAL} & \textit{22.36} & \textit{16.90} & \textit{22.29} \\
    \multicolumn{5}{l}{\textit{Nuggets (response independent)}} \\
    & vital & \multicolumn{3}{c}{3.13} \\
    & okay & \multicolumn{3}{c}{9.00} \\
    & \textit{TOTAL} & \multicolumn{3}{c}{\textit{12.13}} \\
    \hline
  \end{tabular}
  \caption{Counts for subclaims and nuggets averaged across open-ended and single-answer query data. Note that nuggets are generated from source documents, independent of response, and are thus shared across normal, missing, and wrong responses. The label ``less'' denotes less important subclaims.}
  \label{tab:results-counts}
\end{table}


\subsection{Experiments}
We evaluate \ourdataset{} using \factscore{}, \nuggetrecall{}, and \ourmetric{} outlined in section \ref{sec:metrics}. For each metric, we have an evaluation on normal responses, responses missing key information, and responses with wrong key information. We are evaluating factuality systems and have no reason to believe a different LLM would yield a different result, and for this reason we pick \texttt{gpt-4o} \citep{openai2024gpt4ocard} as a representative model of the types of models used for evaluation. See Appendix \ref{app:prompts} for more details and prompts.
We can then compare how metrics change between these three response types, if metrics are sensitive to these minimal errors, and if they are able to detect these errors.

\section{Results}



From Table \ref{tab:results-counts}, we first observe there are fewer nuggets than subclaims (because a nugget often contains multiple subclaims) when comparing the same subset (e.g., normal open-ended subclaim total vs. nuggets total), and fewer subclaims/nuggets for single-answer queries than open-ended queries, specifically vital ones. 
Responses with missing key information are shorter, and therefore contain fewer subclaims, while responses with wrong key information still have a similar number of subclaims as the original response.

\begin{table*}[ht]
  \centering
  \begin{tabular}{cllccc}
    \hline
    \textbf{Metric Type} & \textbf{Dataset} & \textbf{Metric} & \textbf{Normal (\%)} & \textbf{Missing (\%)} & \textbf{Wrong (\%)} \\
    \hline
    \multirow{4}{*}{Precision} & \multirow{2}{*}{Open-Ended} & \factscore{} & 83.62 & 83.58 & 78.84 \\
     & & \vitalprecision{} & 82.69 & 82.88 & 75.40 \\
    \cline{2-6}
     & \multirow{2}{*}{Single-Answer} & \factscore{} & 82.58 & 82.75 & 76.63 \\
     & & \vitalprecision{} & 72.81 & 72.22 & 48.73 \\
    \hline
    \multirow{4}{*}{Recall} & \multirow{2}{*}{Open-Ended} & \nuggetrecall{} & 24.35 & 18.61 & 23.23 \\
     & & \vitalrecall{} & 40.90 & 29.91 & 37.89 \\
    \cline{2-6}
     & \multirow{2}{*}{Single-Answer} & \nuggetrecall{} & 27.71 & 19.13 & 23.49 \\
     & & \vitalrecall{} & 52.44 & 27.45 & 36.70 \\
    \hline
  \end{tabular}
  \caption{Precision and recall results for open-ended and single-answer queries. \vitalprecision{} and \vitalrecall{} show a clear difference between normal and adversarial responses, where \factscore{} and \nuggetrecall{} are less sensitive.}
  \label{tab:results-precision-recall}
\end{table*}

There are similar trends in precision and recall metrics (Table \ref{tab:results-precision-recall}). 
In \factscore{} results, we observe normal and missing responses are about the same, with only a small dip (about a 5-7\% dip) for wrong responses. For open-ended query \vitalprecision{}, we see a similar decrease. In contrast, we observe a large drop between normal/missing and wrong responses in \vitalprecision{} (about 25\% less) for single-answer queries. For \nuggetrecall{}, we observe all responses (normal/missing/wrong) are within 6\% of one another for open-ended responses and 9\% for single-answer responses, with the highest score for normal responses, and lowest for missing responses. Open-ended \vitalrecall{} exhibits similar trends, with an 11\% difference for normal and missing responses. We see a much larger dip of 25\% for single-answer queries.

Response-level metrics in Table \ref{tab:results-errors} are the best indicator of these errors in key information. Note that here, lower numbers are better, as they indicate a lower percent of error.
\vitalprecision{} reports the percentage of responses with any vital subclaims wrong. We observe good detection of incorrect responses here, with 75\% (open-ended) and almost 90\% (single-answer) of wrong responses getting vital subclaims wrong.
\vitalrecall{} indicates the percentage of responses where any vital nugget is unsupported. There is a small difference in \vitalrecall{} scores for all responses (less than a 5\% difference) for open-ended queries. However, \vitalrecall{} is a better detector of error in single-answer queries, with a 15\% difference between normal responses and missing/wrong responses.

\begin{table}[b]
  \centering
  \begin{tabular}{llrrr}
    \hline
    \textbf{     } & \textbf{Metric} & \textbf{Normal} & \textbf{Missing} & \textbf{Wrong} \\
    \hline
    \multicolumn{5}{l}{\textbf{Open-Ended}} \\
    & \responselevelprecision{} & 52.70 & 45.48 & 73.79 \\
    & \responselevelrecall{} & 87.22 & 91.18 & 88.58 \\
    \hline
    \multicolumn{5}{l}{\textbf{Single-Answer}} \\
    & \responselevelprecision{}  & 45.03 & 42.17 & 89.53 \\
    & \responselevelrecall{} & 54.30 & 68.10 & 68.47 \\
    \hline
    \end{tabular}
  \caption{Response level metric results best capture adversarial edits. Wrong or missing vital claims are detected in adversarial responses.}
  \label{tab:results-errors}
\end{table}




\section{Analysis}

\subsection{\ourmetric{} Metrics}

\paragraph{Metric changes}
Our experiments demonstrate that \ourmetric{} metrics are more sensitive to key information errors than prior methods. Response-level \ourmetric{} metrics in particular are best at differentiating response quality for responses with missing and wrong key information, especially for responses to single-answer queries, where missing or wrong key information directly impacts overall response correctness.
Overall there is less of a difference in normal/missing response scores, indicating that it is more difficult to detect omissions than it is to identify wrong information.
A crucial piece in this metric is the ranking and importance labeling of subclaims. Unlike \factscore{}, which equally distributes weight across subclaims, \ourmetric{} metrics highlight which claims are most critical for answering a query. \ourmetric{} metrics' importance weighting better 
reflects user perception: mistakes in vital information undermine trust more than errors in peripheral details.

\paragraph{Subclaim ranking}
An crucial step in the precision evaluation is the subclaim ranking and importance labeling. An example of subclaim ranking is shown in Table \ref{tab:rank-example}. The first decomposed subclaim in a normal response is ``X is a song,'' which does not immediately answer the query, and is labeled as less important. After ranking, there are only two vital subclaims, the two which answer the query. Most of the subclaims are labeled as less important background knowledge, not needed for answering the query. The second response is missing key information. Because key information is missing, there are no subclaims labeled as vital. The last response contains the same first subclaim as the normal response, which also is labeled as less important. Crucially, despite the answers in the subclaims being wrong, they are labeled as vital subclaims because they are answering exactly what the query is asking.

\subsection{\ourdataset{}}

\paragraph{Adversarial responses}
An example instance is shown in Table \ref{tab:data-examples} for the query ``Of which tribe of Red Indians was Geronimo a chief?'' The correct answer to this query is the Apache tribe, which is the written in the \textit{normal} response. The \textit{missing} response omits the first sentence of the normal response, while the \textit{wrong} response changes ``Apache'' to ``Sioux.'' Later mentions of the Apache tribe still remain in the \textit{wrong} response, however the initial answer is incorrect. The initial answer to the query is the part of the answer which LLM users will pay the most attention to. The \textit{wrong} adversarial data may contain contradictions, however the initial answer is altered. 

\paragraph{Cumulative precision} To assess how key information errors can get buried in a long response, we plot the cumulative \factscore{} for the first $n$ subclaims. This cumulative \factscore{} for single-answer queries plot can be found in Figure \ref{fig:cumulative-sa}. We observe the same trend with open-ended queries, shown in the appendix in Figure \ref{fig:cumulative-oe}. In the adversarial data with wrong key information, we observe a lower averaged \factscore{} of around 60\% at the start of responses. In the data, the first subclaim is oftentimes an obvious statement used as exposition for the answer (e.g., first subclaim is ``X is a song.'' in Table \ref{tab:rank-example}), which explains the slight dip in precision between the first two subclaims. The normal response and the response missing key information has a high cumulative \factscore{} across all subclaims. 

\paragraph{Single-answer vs open-ended queries} In single-answer queries, there is one answer the query is looking for (e.g., ``Who is the author of \textit{The Great Gatsby}?''). In open-ended queries the answer is less defined, and there are many possible valid answers (e.g., ``Tell me a bio of F. Scott Fitzgerald'').

\begin{figure}[!t]
    \centering
    \includegraphics[width=1\linewidth]{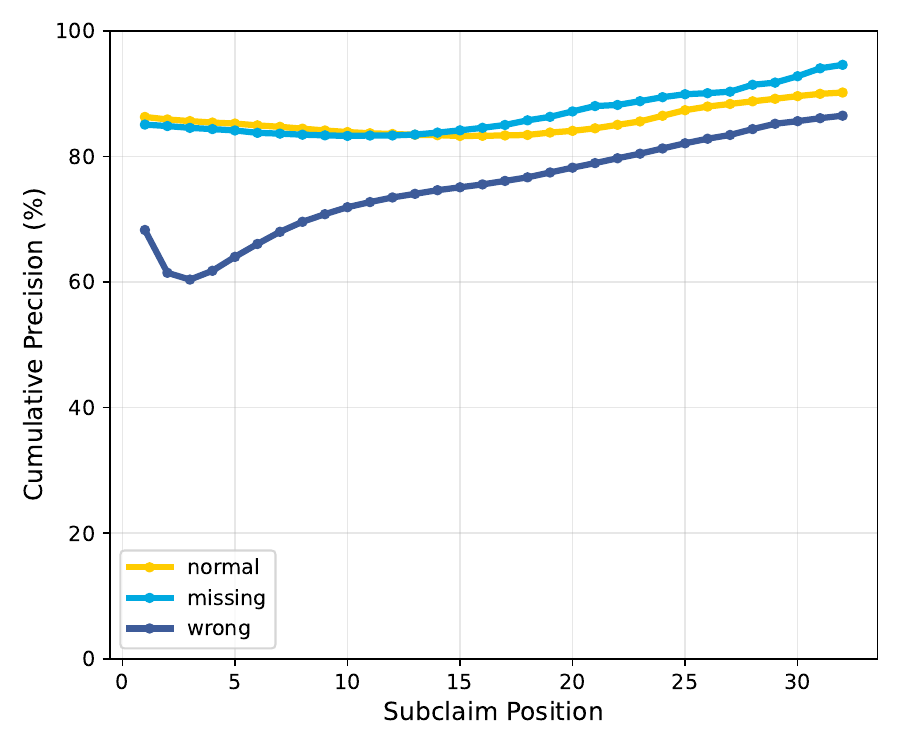}
    \caption{Cumulative precision over subclaim position for single-answer queries. Wrong responses show low precision early on due to falsified key claims, while normal and missing responses stay consistently high.}
    \label{fig:cumulative-sa}
\end{figure}

The evaluation of open-ended queries with nuggets is difficult. Nugget creation is dependent on what documents are provided during the nugget creation process, and therefore to get a high recall score, the generated text should match the content of the provided documents, which is not always likely to happen, especially for entities with diverse masses of information. For these open-ended queries, it is not clear how to determine what the most important piece of information is. For example, in a human biography, it is not established if a person's occupation, accomplishments, birth place, etc. is more important than the other. All of these pieces of information tend to be included in a biography. Because of this, there are far more nuggets for these queries than for single-answer queries. In the response-level metrics, this means there is more opportunity to make a mistake in one of the vital subclaims, explaining the high percentage of responses with any vital nuggets unsupported. On the other hand, using nuggets to evaluate single-answer queries is quite simple. The answer need only contain a few pieces of information, which make up the nuggets. Fewer nuggets means there is less opportunity to make a mistake.

These issues are similar in subclaim importance labeling and ranking. Responses to open-ended queries may cover a lot of ground, with an unclear hierarchy to the subclaims within the response. This results in far more claims labeled as vital, with more possibilities of getting at least one vital subclaim wrong.

\section{Conclusion}

In this paper, we demonstrated how existing factual precision methods fail to distinguish between errors in key information. We introduced \ourmetric{}, a set of metrics designed to evaluate key information factual accuracy and recall. Through the construction of \ourdataset{} dataset, we showed that current methods are insensitive and assign scores to responses of drastically different quality. \ourmetric{} metrics show moderate improvements over baseline precision and recall, with the response-level evaluations best capturing the impact of key information errors. 

These results suggest that any robust factuality evaluation should explicitly incorporate importance-weighted subclaims rather than treating all information as equally important.
One way of doing this is by reporting our \ourmetric{} decomposition- or response-level metrics to evaluate only the most critical information in responses.
Another way to do this is to weight subclaims by their importance, using our prompt-based importance ranking. However, it is not entirely clear where the weight values come from. We leave this as an open question, which would require alignment with user preference, and is potentially dependent on query type or topic. If each subclaim is strictly more important than a subclaim ranked lower, then linear decay weighting is an option (see Appendix \ref{appendix:results}), but this is a strong assumption and not applicable in most cases.
We recommend the evaluation of new factuality metrics on \ourdataset{} to see if insensitivities exist, and if they are able to distinguish errors in difficult responses.
Together, \ourmetric{} metrics and \ourdataset{} offer a new benchmark in factual precision for evaluating and detecting errors in key information.

\section*{Limitations}

In this study, we demonstrate shortcomings of factual precision metrics, and introduce \ourmetric{} and \ourdataset{} to solve and evaluate these issues. Our approach depends on LLMs for many aspects of the framework, including decomposition, ranking, and verification. The method inherits bias and inconsistencies from the underlying model.

Like other factuality metrics, \ourmetric{} is dependent on the quality of the retrieved reference material used for verification. In this work, we use sources already paired with the query from previous datasets, and assume this grounding is factual and relevant, but using this framework with a retriever model adds another variable.
Poor grounding can cause false positives or false negatives, a challenge that is particularly difficult in domains where high-quality references are sparse.

\ourmetric{} introduces an additional step over standard precision and recall frameworks, adding a ranking/importance labeling step. This step improves metric sensitivity, but requires more LLM calls (an extra importance ranking call per response), which may limit scalability. Despite these limitations, \ourmetric{} and \ourdataset{} are important steps towards evaluation methods that reflect the \textit{importance} of factual errors.


\section*{Ethics Statement}

This work addresses the evaluation of factuality in LLM outputs, an area with direct implications to safe and trustworthy deployment of these systems. By introducing \ourmetric{} and \ourdataset{}, we aim to promote evaluation practices that more accurately identify harmful or misleading responses. Our dataset \ourdataset{} was constructed using automated adversarial perturbations of machine-generated text, which ensures scalability, but may not capture the full diversity of real-world factual errors. Importance judgments for subclaims and nuggets are automatic by LLMs, which introduces the risk of reinforcing biases present in the model, either under- or over-weighting certain parts of a response, depending on what the model views as ``important.'' Although our goal is to improve factuality evaluation, any benchmark can be misused if it is treated as a definite or exhaustive measure of reliability. We caution against over-reliance in high-stakes contexts (e.g., medical), without additional safeguards, domain-specific evaluation, and human oversight.

\bibliography{custom}
\appendix
\clearpage

\section{Prompts, Model Details, Compute}\label{app:prompts}

The prompts used for generation of normal responses are shown in Table \ref{tab:normal-prompts}, and adversarial responses are shown in Table \ref{tab:adversarial-prompts}. For normal response generation, we prepend an instruction to answer in a paragraph for each of the queries, to ensure the text is in long-form. For missing and wrong response generation, we provide the LLM with the query and normal LLM response, and instruct the LLM to adversarially omit or falsify key information in the response. For both normal and adversarial response generation, we use \texttt{gpt-4o} \citep{openai2024gpt4ocard}, with a temperature of 0.2 and 2000 max tokens.

\begin{table*}[ht]
\centering
\begin{tabular}{m{2.4cm}m{2.5cm}m{8.5cm}}
\toprule
\textbf{Data Type} & \textbf{Dataset} & \textbf{Prompt} \\ 
\hline
\multirow{5}{*}{Open-Ended} & \factscore{} Bios & In a paragraph, tell me a bio of [TOPIC] \\ 
\cline{2-3}
& \textsc{WildHal.} & In a paragraph, could you tell me what you know about [TOPIC]? \\
\cline{2-3}
& \bright{} & Answer the following question in a short paragraph: [QUERY]\\
\hline
\multirow{3}{*}{Single-Answer} & \hotpotqa{} & In a paragraph, could you answer: [QUERY]\\
\cline{2-3}
& \naturalquestions{} & In a paragraph, could you answer: [QUERY]\\
\cline{2-3}
& \triviaqa{} & In a paragraph, could you answer: [QUERY]\\
\bottomrule
\end{tabular}
\caption{Prompts used to generate normal long-form LLM responses across datasets, with variations for open-ended and single-answer queries.}\label{tab:normal-prompts}
\end{table*}

\begin{table*}[ht]
\centering
\small
\begin{tabular}{p{1.2cm}p{14cm}}
\toprule
\textbf{Adversary Type} & \textbf{Prompt} \\
\midrule
\multirow{25}{*}{\textbf{Missing}} & You are given a query and an answer. Your task is to modify the answer by removing the most important information for answering the question. You are first given an example, and then your task.\\
& \\
& EXAMPLE\\
& Query: Which is denser, water vapor or air? \\
& Answer: Air is denser than water vapor. This is because the molecular weight of the primary components of air mainly nitrogen and oxygen is higher than that of water vapor. Nitrogen has a molecular weight of approximately 28 g/mol, and oxygen about 32 g/mol, whereas water vapor has a molecular weight of only about 18 g/mol. When water vapor is added to air, it actually displaces some of the heavier nitrogen and oxygen molecules, making the resulting humid air less dense than dry air. This principle is why humid air tends to rise and is a key factor in weather patterns and cloud formation. At the same temperature and pressure, a volume of moist air will weigh less than the same volume of dry air, confirming that water vapor is less dense than the air it mixes with.\\
& Modified answer: This is because the molecular weight of the primary components of air mainly nitrogen and oxygen is higher than that of water vapor. Nitrogen has a molecular weight of approximately 28 g/mol, and oxygen about 32 g/mol, whereas water vapor has a molecular weight of only about 18 g/mol. When water vapor is added to air, it actually displaces some of the heavier nitrogen and oxygen molecules, making the resulting humid air less dense than dry air. This principle is why humid air tends to rise and is a key factor in weather patterns and cloud formation. At the same temperature and pressure, a volume of moist air will weigh less than the same volume of dry air, confirming that water vapor is less dense than the air it mixes with.\\
& \\
& \\
& YOUR TASK\\
& Query: [QUERY]\\
& Answer: [ANSWER]\\
& Modified answer:\\
\midrule
\multirow{28}{*}{\textbf{Wrong}} & You are given a query and its corresponding answer. Your task is to make a modification to one sentence from the answer by changing the key piece of information required to answer the question correctly, thereby making the answer factually incorrect. This will most likely be a change to the first sentence. Do not alter the remainder of the response. An example is provided first, followed by your task. \\
& \\
& EXAMPLE \\
& Query: Which is denser, water vapor or air? \\
& Answer: Air is denser than water vapor. This is because the molecular weight of the primary components of air mainly nitrogen and oxygen is higher than that of water vapor. Nitrogen has a molecular weight of approximately 28 g/mol, and oxygen about 32 g/mol, whereas water vapor has a molecular weight of only about 18 g/mol. When water vapor is added to air, it actually displaces some of the heavier nitrogen and oxygen molecules, making the resulting humid air less dense than dry air. This principle is why humid air tends to rise and is a key factor in weather patterns and cloud formation. At the same temperature and pressure, a volume of moist air will weigh less than the same volume of dry air, confirming that water vapor is less dense than the air it mixes with. \\
& Modified answer: Water vapor is denser than air. This is because the molecular weight of the primary components of air mainly nitrogen and oxygen is higher than that of water vapor. Nitrogen has a molecular weight of approximately 28 g/mol, and oxygen about 32 g/mol, whereas water vapor has a molecular weight of only about 18 g/mol. When water vapor is added to air, it actually displaces some of the heavier nitrogen and oxygen molecules, making the resulting humid air less dense than dry air. This principle is why humid air tends to rise and is a key factor in weather patterns and cloud formation. At the same temperature and pressure, a volume of moist air will weigh less than the same volume of dry air, confirming that water vapor is less dense than the air it mixes with. \\
& \\
& \\
& YOUR TASK \\
& Query: [QUERY] \\
& Answer: [ANSWER] \\
& Modified answer: \\
\bottomrule
\end{tabular}
\caption{Prompts for adversarial data generation. ``Missing'' prompts remove key information, while ``wrong'' prompts replace with incorrect information.}
\label{tab:adversarial-prompts}
\end{table*}

The prompt for importance ranking are shown in Table \ref{tab:importance-rank-prompt}, with an example ranking in Table \ref{tab:rank-example}. The prompt provides an LLM with a query and set of subclaims, asking the model to order subclaims based on importance for answering the query, independent of correctness, and label each subclaim with either \textit{vital}, \textit{okay}, or \textit{less important} importance labels. For the importance ranking, we use \texttt{gpt-4o} with a temperature of 0.2 and 4000 max tokens. In total, generating responses took 15 GPU-hours, evaluating with \factscore{}/nuggets/\ourmetric{} took 450 GPU-hours.

\begin{table*}[ht]
\centering
\small
\begin{tabular}{p{14cm}}
\toprule
You are performing step two of a four part fact-checking process: \\
(1) Decompose a paragraph into individual claims.\\
(2) Given a query and set of claims, rank by decreasing query-importance (this step).\\
(3) Check the correctness of each claim.\\
(4) Score the paragraph, weighting by importance. \\
This step is completely independent of factual correctness, and only focuses on the query-importance of claims for answering the query. Even factually incorrect claims should be ranked highly if they directly answer the query.\\
\\
Instructions: You are provided with a query and set of claims. Rank the claims in decreasing order of query-importance. A claim exhibits high query-importance when it addresses a central aspect of the query, and low query-importance when it contributes only peripheral or background information. Rank claims independent of correctness, instead only based on query-importance. A later step will check for correctness of claims.\\
\\
Assign query-importance labels using exactly these three categories:\\
- "vital" - Essential claims that directly address the core query\\
- "okay" - Supporting claims that provide useful but non-essential information\\
- "less-important" - Background or tangentially related claims with minimal relevance\\
\\
Ordering Rules:\\
- All "vital" claims must appear first, then all "okay" claims come second, and "less-important" claims come last.\\
- Within each category, order by decreasing importance.\\
- If two or more claims address the same aspect of the query, keep them grouped in the order they appear, even if their answers contradict. For example:\\
\quad...\\
\quad$[$S3$]$ Washington, D.C. is the capital of Canada.: "vital"\\
\quad$[$S8$]$ Washington, D.C. is the capital of the United States.: "vital"\\
\quad...\\
- Do not adjust rankings based on factual correctness, this will be handled in step 3.\\
\\
Output Format:\\
$[$Claim ID$]$ <claim text>: "label"\\
$[$Claim ID$]$ <claim text>: "label"\\
...\\
\\
Requirements:\\
- Label every claim exactly once\\
- Use only the three specified labels\\
- Maintain the original claim count\\
- Return only the labeled, ordered list (no explanations)\\
Below is your task.\\
\\
\#\#\#Your task:\\
Query: [QUERY]\\
Claims:\\
$[$SUBCLAIMS$]$\\
Ranked Claims:\\
\bottomrule
\end{tabular}
\caption{Prompt for ranking subclaims by \textit{query importance} and importance labeling. The LLM assigns each subclaim with a ``vital'', ``okay'', or ``less-important'' label independent of corectness.}
\label{tab:importance-rank-prompt}
\end{table*}

\begin{table*}[ht]
\centering
\small
\begin{tabular}{p{1cm}p{0.5cm}p{1.9cm}p{11cm}}
\toprule
\multirow{1}{*}{\textbf{Query}} & \multicolumn{3}{c}{In a paragraph, could you answer: who wrote i want to dance with somebody by whitney houston} \\
\midrule
\textbf{Type} & \textbf{Index} & \textbf{Importance} & \textbf{Response} \\
\midrule
\multirow{23}{*}{\textbf{Normal}} & 2 & vital & "I Wanna Dance with Somebody (Who Loves Me)" was written by George Merrill.\\
& 3 & vital & "I Wanna Dance with Somebody (Who Loves Me)" was written by Shannon Rubicam.\\
& 7 & okay & The duo wrote (penned) the song specifically for Whitney Houston.\\
& 4 & okay & George Merrill is a songwriter.\\
& 5 & okay & Shannon Rubicam is a songwriter.\\
& 6 & okay & The duo performed together as the pop group Boy Meets Girl.\\
& 1 & less-important & "I Wanna Dance with Somebody (Who Loves Me)" is a song.\\
& 8 & less-important & Boy Meets Girl is a pop group.\\
& 9 & less-important & Whitney Houston is a singer (artist).\\
& 10 & less-important & The track was released in 1987.\\
& 11 & less-important & The track was the lead single from her second studio album.\\
& 12 & less-important & Her second studio album is titled "Whitney.".\\
& 13 & less-important & The track is from the album "Whitney.".\\
& 14 & less-important & The track became one of Houston's most iconic hits.\\
& 15 & less-important & The track was released as the lead single from "Whitney" in 1987.\\
& 16 & less-important & The song has an upbeat tempo.\\
& 17 & less-important & The song has an infectious melody.\\
& 18 & less-important & The song features Houston's powerful vocals.\\
& 19 & less-important & The combination of the song's upbeat tempo, infectious melody, and Houston's powerful vocals contributed to its success.\\
& 20 & less-important & The song reached the top of the charts worldwide.\\
& 21 & less-important & The success of the song helped solidify Houston's status as a global pop superstar.\\
& 22 & less-important & Houston is a global pop superstar.\\
\midrule
\multirow{14}{*}{\textbf{Missing}} & 1 & okay & The track was released in 1987.\\
& 2 & okay & The track was the lead single from her second studio album.\\
& 4 & okay & The track is from the album "Whitney.".\\
& 3 & okay & Her second studio album is titled "Whitney.".\\
& 5 & okay & The track became one of Houston's most iconic hits.\\
& 11 & okay & The song reached the top of the charts worldwide.\\
& 12 & okay & The song helped solidify Houston's status as a global pop superstar.\\
& 6 & less-important & Houston has multiple iconic hits.\\
& 7 & less-important & The song has an upbeat tempo.\\
& 8 & less-important & The song has an infectious melody.\\
& 9 & less-important & The song features Houston's powerful vocals.\\
& 10 & less-important & The upbeat tempo, infectious melody, and Houston's powerful vocals helped propel the song to the top of the charts worldwide.\\
& 13 & less-important & Houston is a global pop superstar.\\
\midrule
\multirow{24}{*}{\textbf{Wrong}} & 2 & vital & "I Wanna Dance with Somebody (Who Loves Me)" was written by Diane Warren.\\
& 3 & vital & "I Wanna Dance with Somebody (Who Loves Me)" was written by David Foster.\\
& 8 & vital & The duo penned the song.\\
& 9 & vital & The duo penned the song specifically for Whitney Houston.\\
& 4 & okay & Diane Warren is a songwriter.\\
& 5 & okay & David Foster is a songwriter.\\
& 6 & okay & The duo performed together.\\
& 7 & okay & The duo performed together as the pop group Boy Meets Girl.\\
& 10 & less-important & Boy Meets Girl is a pop group.\\
& 1 & less-important & "I Wanna Dance with Somebody (Who Loves Me)" is a song.\\
& 11 & less-important & Whitney Houston is a singer.\\
& 12 & less-important & The track was released in 1987.\\
& 13 & less-important & The track was the lead single from her second studio album.\\
& 14 & less-important & Her second studio album is titled "Whitney.".\\
& 15 & less-important & The track was released as the lead single from the album "Whitney.".\\
& 16 & less-important & The track became one of Houston's most iconic hits.\\
& 17 & less-important & Houston had a second studio album titled "Whitney.".\\
& 18 & less-important & The song has an upbeat tempo.\\
& 19 & less-important & The song has an infectious melody.\\
& 20 & less-important & The song features Houston's powerful vocals.\\
& 21 & less-important & The combination of the song's upbeat tempo, infectious melody, and Houston's powerful vocals contributed to its success.\\
& 22 & less-important & The song reached the top of the charts worldwide.\\
& 23 & less-important & The song's success helped solidify Houston's status as a global pop superstar.\\
\bottomrule
\end{tabular}
\caption{Example subclaim rankings for a query from \naturalquestions{}. Normal responses highlight correct vital subclaims, while wrong responses show incorrect, but still vital subclaims.}
\label{tab:rank-example}
\end{table*}

\section{Results}\label{appendix:results}

\subsection{Cumulative Precision}
\begin{figure}[!h]
    \centering
    \includegraphics[width=1\linewidth]{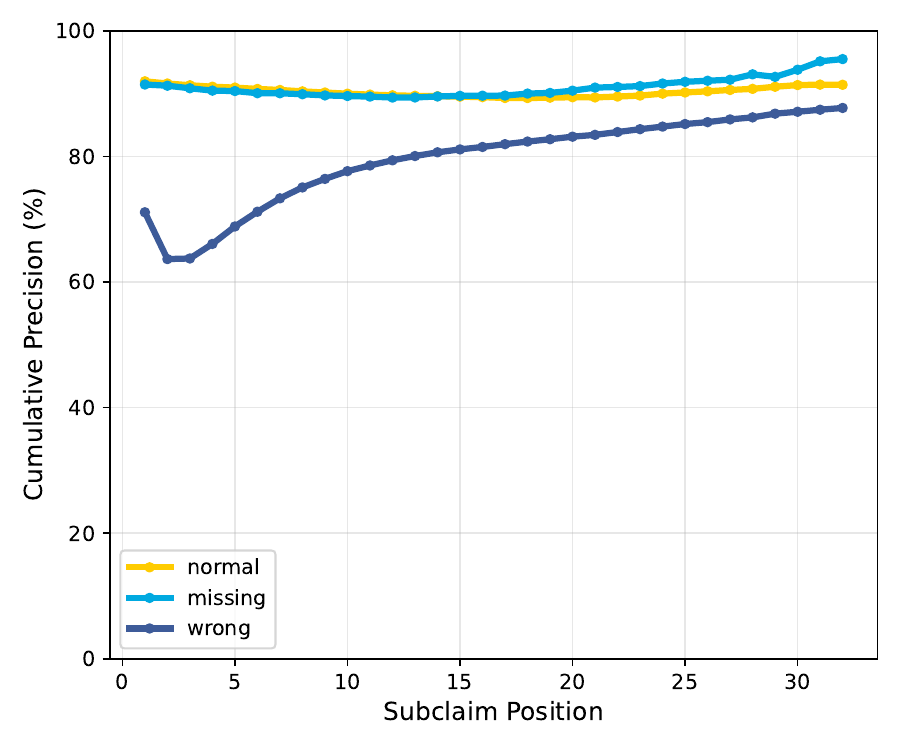}
    \caption{Cumulative precision over subclaim position for open-ended queries. Wrong responses show low precision early on due to falsified key claims, while normal and missing responses stay consistently high.}
    \label{fig:cumulative-oe}
\end{figure}

The cumulative \factscore{} precision for open-ended queries is shown in Figure \ref{fig:cumulative-oe}. This demonstrates how key information errors get hidden in long responses. Similarly to single-answer queries (Figure \ref{fig:cumulative-sa}), we observe responses with wrong key information receive low initial scores, which increases with response length.

\subsection{Linear Decay Weighting}
We report the precision and recall with a linear decay weighting. This weighting assumes that each subclaim/nugget ranked higher than another is strictly more important. This is an alternative version of the \ourmetric{} metrics, however the assumption that each subclaim/nugget is strictly more important than the ones ranked below is a strong assumption that is not usually true. For this reason, we do not include it in \ourmetric{} metrics, but include it here to demonstrate performance under these assumptions.

\begin{figure*}
    \centering
    \includegraphics[width=1\linewidth]{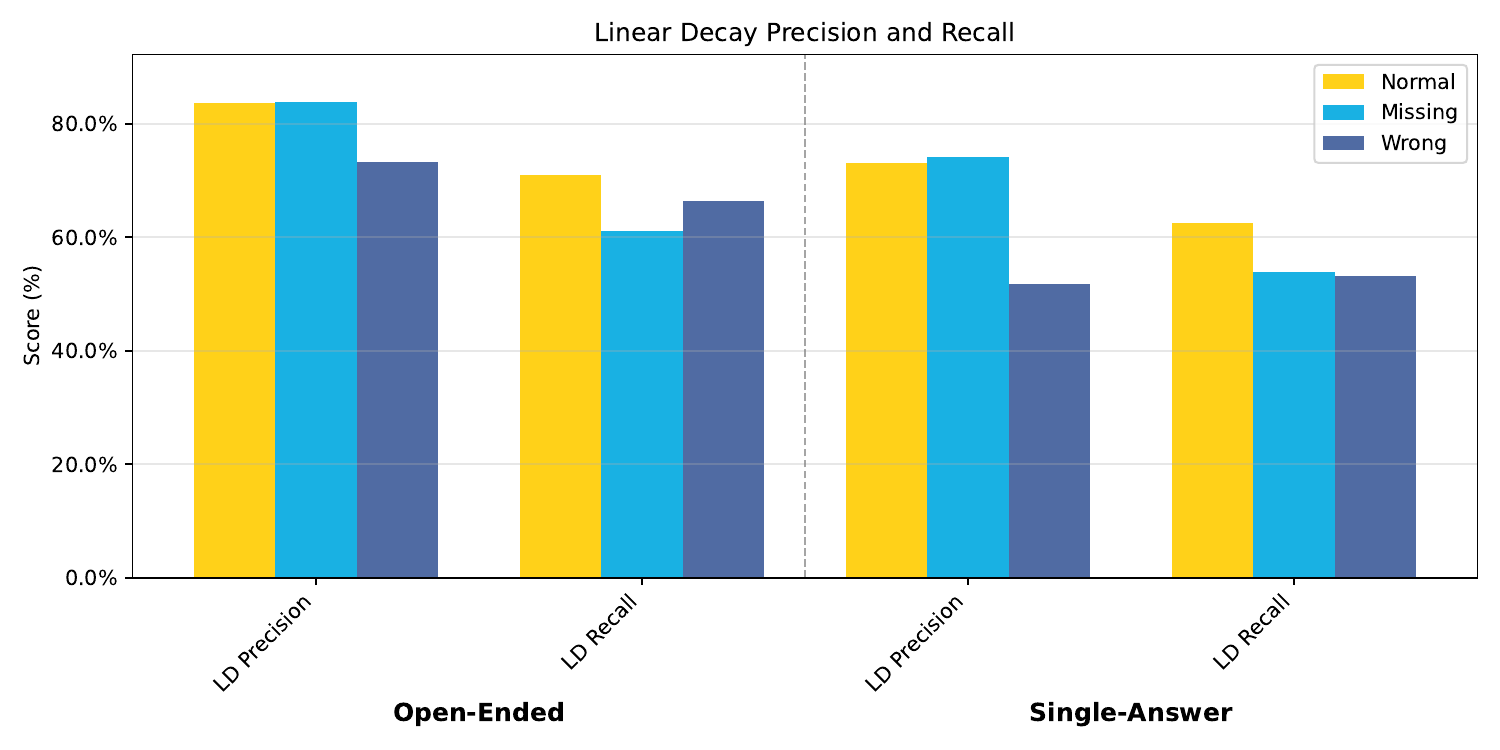}
    \caption{Linear decay weighted precision and recall. Highlights the differences between normal, missing, and wrong responses, though less strongly than response-level metrics.}
    \label{fig:linear-decay}
\end{figure*}

\end{document}